\documentclass[10pt]{article}

\usepackage[utf8]{inputenc}
\usepackage[T1]{fontenc}
\usepackage{lmodern}
\usepackage[a4paper,margin=1in]{geometry}
\usepackage{amsmath,amssymb}
\usepackage[numbers,sort&compress]{natbib}
\usepackage{xcolor}
\usepackage{hyperref}
\hypersetup{colorlinks=true,linkcolor=blue,citecolor=blue,urlcolor=blue}

\usepackage{caption}
\newcommand{\copyrightyear}[1]{}
\newcommand{\copyrightclause}[1]{}
\newcommand{\conference}[1]{}
\newcommand{\cormark}[1]{}
\newcommand{\cortext}[2]{}
\newcommand{\sep}{,\ }
\newenvironment{keywords}{\par\noindent\textbf{Keywords}\par\noindent}{\par}
\newenvironment{acknowledgments}{\section*{Acknowledgments}}{}

\usepackage{listings}
\usepackage{algorithm}
\usepackage{algpseudocode}
\usepackage{tikz}
\usetikzlibrary{
  arrows.meta,
  positioning,
  calc,
  shadows.blur,
  shapes.geometric,
  shapes.symbols
}
\usepackage{booktabs}
\usepackage{graphicx}
\makeatletter
\let\EFL@origincludegraphics\includegraphics
\renewcommand{\includegraphics}[2][]{%
  \IfFileExists{#2}{\EFL@origincludegraphics[#1]{#2}}{%
    \fbox{\parbox[c][0.09\textheight][c]{0.8\linewidth}{\centering\scriptsize Missing figure file: \texttt{\detokenize{#2}}}}%
  }%
}
\makeatother

\DeclareMathAlphabet{\mathcal}{OMS}{cmsy}{m}{n}
\begin{document}

\copyrightyear{2026}
\copyrightclause{Copyright for this paper by its authors.
  Use permitted under Creative Commons License Attribution 4.0
  International (CC BY 4.0).}

\conference{The Annual Symposium of Computer Science 2026}

\title{Entropy-Regularized Probabilistic Gates for Sparse Model Discovery in Scarce-Data Federated Learning}

\author{
Krishna Harsha Kovelakuntla Huthasana \\
Alireza Olama \quad Andreas Lundell \\[0.5em]
Department of Engineering and Information Technology, Åbo Akademi University \\
\texttt{\{kkovelak, alireza.olama, andreas.lundell\}@abo.fi}
}
\maketitle

\begin{abstract}
\noindent Federated Learning (FL) is a distributed machine learning (ML) paradigm with collaboration among multiple clients without sharing data. FL is challenging under data heterogeneity and partial client participation. Learning sparse models is useful for communication and computational efficiency in FL, but it is especially difficult in the small-sample high-dimensional regime $(d \gg N)$ where optimization can yield parameter configurations that fail to generalize to unseen test data. While magnitude-based pruning doesn't account for uncertainty exploration in the parameter space, a formulation with probabilistic gates and an $L_{0}$ constraint allows sampling from competing sparse configurations during training. In this work, we study entropy regularization of gate distributions as a mechanism to maintain uncertainty in sparse federated optimization by preventing early commitment to sparse support. We examine its impact under data heterogeneity, client participation heterogeneity, and sparsity. Experiments on synthetic and real-world benchmarks show consistent improvements over federated iterative hard thresholding (Fed-IHT) and pruning after dense federated averaging (FedAvg) training, both in statistical performance on test data and in sparsity recovery accuracy.
\end{abstract}

\section{Introduction}
\begin{figure}[htbp]
\centering
\resizebox{0.7\linewidth}{!}{%
\begin{tikzpicture}[
  font=\sffamily,
  arrowblue/.style={
    -{Triangle[length=5mm,width=4mm]},
    line width=1.5pt,
    draw=blue
  },
  greenflow/.style={
    -{Triangle[length=4mm,width=3.5mm]},
    dashed,
    line width=1.4pt,
    draw=green!70!black
  },
  serverunit/.style={
    draw=black,
    rounded corners=2pt,
    line width=1pt,
    fill=blue!35!black,
    minimum width=2.8cm,
    minimum height=0.55cm
  }
]

\node[
  draw=blue,
  rounded corners=8pt,
  line width=1pt,
  minimum width=7.4cm,
  minimum height=7cm
] (serverbox) at (0,3.75) {};

\node[font=\bfseries\huge, text=blue!30!black] at (0,5.55) {Server};

\foreach \y in {4.75,4.10,3.45} {
  \node[serverunit] at (0,\y) {};
  \fill[green!90!black] (-1.15,\y) circle (0.07);
  \fill[cyan] (-0.88,\y) circle (0.07);

  \foreach \x in {0.52,0.72,0.92,1.12} {
    \draw[black,line width=1pt] (\x,\y+0.13) -- (\x,\y-0.13);
  }

  \foreach \yy in {-0.10,0,0.10} {
    \draw[black,line width=1pt] (0.45,\y+\yy) -- (1.25,\y+\yy);
  }
}

\draw[fill=blue!15, draw=none, opacity=0.5] (0,3.10) ellipse (1.55cm and 0.12cm);

\draw[greenflow] (-2.8,0.55) -- (-1.25,2.45);
\draw[greenflow] (-1.1,0.50) -- (-0.45,2.35);
\draw[greenflow] (1.1,0.50) -- (0.45,2.35);
\draw[greenflow] (2.8,0.55) -- (1.25,2.45);

\draw[blue,line width=1.5pt] (0,0.20) -- (0,-0.25);
\draw[blue,line width=1.5pt] (-7.2,-0.25) -- (7.2,-0.25);

\foreach \x in {-7.2,-2.7,2.7,7.2} {
  \draw[arrowblue] (\x,-0.25) -- (\x,-1.25);
}

\foreach \x in {-6.2,-1.65,1.75,6.25} {
  \draw[greenflow] (\x,-2.05) -- (\x,-0.30);
}

\begin{scope}[shift={(-7.45,-3.0)}]
  \draw[fill=black, rounded corners=2pt] (-0.75,-0.35) rectangle (0.75,0.80);
  \fill[blue!45] (-0.65,-0.25) rectangle (0.65,0.68);
  \fill[white, opacity=0.18] (0.05,-0.25) -- (0.45,-0.25) -- (0.72,0.68) -- (0.30,0.68) -- cycle;
  \draw[fill=black!85] (-0.95,-0.62) -- (0.95,-0.62) -- (0.62,-0.35) -- (-0.62,-0.35) -- cycle;
  \draw[fill=gray!50] (-0.75,-0.55) -- (0.75,-0.55) -- (0.52,-0.40) -- (-0.52,-0.40) -- cycle;

  \begin{scope}[shift={(1.25,-0.05)}]
    \draw[fill=black!90, rounded corners=3pt] (-0.25,-0.55) rectangle (0.25,0.42);
    \fill[blue!45] (-0.20,-0.40) rectangle (0.20,0.30);
    \fill[white, opacity=0.20] (-0.02,-0.40) -- (0.18,-0.40) -- (0.25,0.30) -- (0.08,0.30) -- cycle;
    \fill[black!30] (0,-0.48) circle (0.04);
  \end{scope}
\end{scope}

\node[font=\bfseries\Large, text=blue!30!black] at (-7.2,-4.35) {Client 1};

\begin{scope}[shift={(-3.0,-3.0)}]
  \draw[fill=black, rounded corners=2pt] (-0.75,-0.35) rectangle (0.75,0.80);
  \fill[blue!45] (-0.65,-0.25) rectangle (0.65,0.68);
  \fill[white, opacity=0.18] (0.05,-0.25) -- (0.45,-0.25) -- (0.72,0.68) -- (0.30,0.68) -- cycle;
  \draw[fill=black!85] (-0.95,-0.62) -- (0.95,-0.62) -- (0.62,-0.35) -- (-0.62,-0.35) -- cycle;
  \draw[fill=gray!50] (-0.75,-0.55) -- (0.75,-0.55) -- (0.52,-0.40) -- (-0.52,-0.40) -- cycle;

  \begin{scope}[shift={(1.45,-0.05)}]
    \draw[fill=white, rounded corners=4pt] (-0.42,-0.25) rectangle (0.42,0.35);
    \draw[fill=white] (-0.28,-0.85) -- (-0.08,-0.25);
    \draw[fill=white] (0.28,-0.85) -- (0.08,-0.25);
    \draw[fill=white] (-0.34,-0.85) -- (0.34,-0.85);
    \fill[black] (0.24,0.05) circle (0.27);
    \fill[cyan!70!blue] (0.24,0.05) circle (0.17);
    \fill[black] (0.24,0.05) circle (0.09);
    \fill[white] (0.15,0.16) circle (0.035);
  \end{scope}
\end{scope}

\node[font=\bfseries\Large, text=blue!30!black] at (-2.7,-4.35) {Client 2};

\begin{scope}[shift={(1.35,-3.0)}]
  \draw[fill=black, rounded corners=2pt] (-0.75,-0.35) rectangle (0.75,0.80);
  \fill[blue!45] (-0.65,-0.25) rectangle (0.65,0.68);
  \fill[white, opacity=0.18] (0.05,-0.25) -- (0.45,-0.25) -- (0.72,0.68) -- (0.30,0.68) -- cycle;
  \draw[fill=black!85] (-0.95,-0.62) -- (0.95,-0.62) -- (0.62,-0.35) -- (-0.62,-0.35) -- cycle;
  \draw[fill=gray!50] (-0.75,-0.55) -- (0.75,-0.55) -- (0.52,-0.40) -- (-0.52,-0.40) -- cycle;

  \begin{scope}[shift={(1.05,-0.05)}]
    \draw[fill=black!90, rounded corners=3pt] (-0.25,-0.55) rectangle (0.25,0.42);
    \fill[blue!45] (-0.20,-0.40) rectangle (0.20,0.30);
    \fill[white, opacity=0.20] (-0.02,-0.40) -- (0.18,-0.40) -- (0.25,0.30) -- (0.08,0.30) -- cycle;
    \fill[black!30] (0,-0.48) circle (0.04);
  \end{scope}

  \begin{scope}[shift={(1.85,-0.08)}]
    \draw[fill=white, rounded corners=4pt] (-0.34,-0.55) rectangle (0.34,0.50);
    \fill[black!85] (0,0.23) circle (0.20);
    \fill[cyan!70!blue] (0,0.23) circle (0.09);
    \fill[black] (0,0.23) circle (0.045);
    \foreach \x in {-0.18,0,0.18} {
      \foreach \y in {-0.32,-0.20,-0.08} {
        \fill[gray!50] (\x,\y) circle (0.018);
      }
    }
  \end{scope}
\end{scope}

\node[font=\bfseries\Large, text=blue!30!black] at (2.2,-4.35) {Client 3};

\begin{scope}[shift={(5.95,-3.0)}]
  \draw[fill=black, rounded corners=2pt] (-0.75,-0.35) rectangle (0.75,0.80);
  \fill[blue!45] (-0.65,-0.25) rectangle (0.65,0.68);
  \fill[white, opacity=0.18] (0.05,-0.25) -- (0.45,-0.25) -- (0.72,0.68) -- (0.30,0.68) -- cycle;
  \draw[fill=black!85] (-0.95,-0.62) -- (0.95,-0.62) -- (0.62,-0.35) -- (-0.62,-0.35) -- cycle;
  \draw[fill=gray!50] (-0.75,-0.55) -- (0.75,-0.55) -- (0.52,-0.40) -- (-0.52,-0.40) -- cycle;

  \begin{scope}[shift={(1.05,-0.05)}]
    \draw[fill=black!90, rounded corners=3pt] (-0.25,-0.55) rectangle (0.25,0.42);
    \fill[blue!45] (-0.20,-0.40) rectangle (0.20,0.30);
    \fill[white, opacity=0.20] (-0.02,-0.40) -- (0.18,-0.40) -- (0.25,0.30) -- (0.08,0.30) -- cycle;
    \fill[black!30] (0,-0.48) circle (0.04);
  \end{scope}

  \begin{scope}[shift={(1.85,-0.08)}]
    \draw[fill=white, rounded corners=3pt] (-0.30,-0.58) rectangle (0.30,0.45);
    \draw[fill=black!85, rounded corners=2pt] (-0.17,0.03) rectangle (0.17,0.34);
    \foreach \y in {0.09,0.18,0.27} {
      \draw[cyan,line width=1pt] (-0.10,\y) -- (0.10,\y);
    }
    \draw[black,line width=0.7pt] (-0.16,-0.20) -- (0.16,-0.20);
  \end{scope}
\end{scope}

\node[font=\bfseries\Large, text=blue!30!black] at (6.9,-4.35) {Client 4};

\end{tikzpicture}%
}
\caption{Client--server federated learning architecture with central orchestration. Solid arrows indicate the aggregated global model distributed to clients, while dotted arrows indicate local model updates sent from clients to the server.}
\label{fig:fl_illustration}
\end{figure}
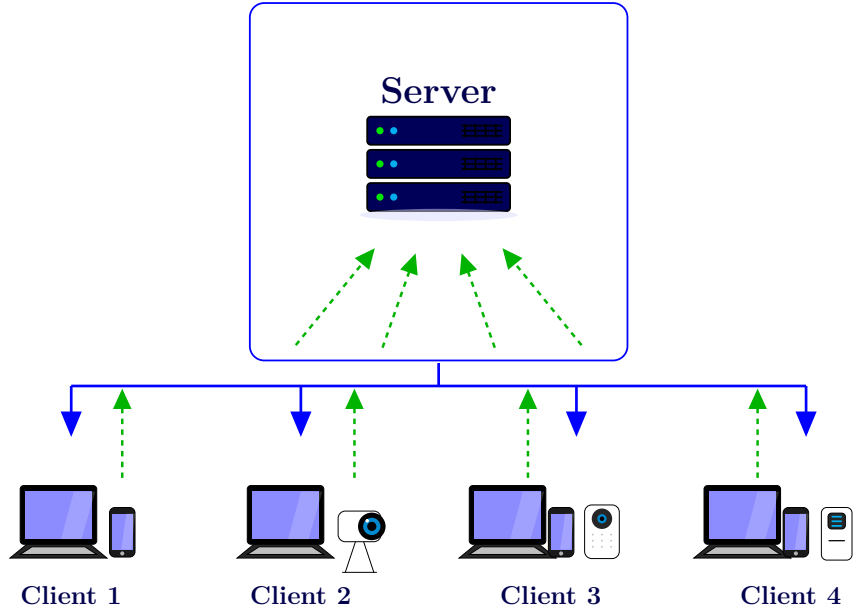

Federated Learning (FL) algorithms operate in a distributed machine learning (ML) setting in which multiple clients collaborate to train \citep{1}. This framework is characterized by privacy requirements of each client and avoids data sharing. While not all distributed settings are privacy-sensitive, FL is still beneficial because it eliminates the need for centralized data \citep{2}. FL can be coordinated either by a single server or by clients communicating with one another. In this work, we study FL with central orchestration by a server to obtain a single global model as shown in Figure  \ref{fig:fl_illustration}.
A global model is typically learned by iteratively averaging parameters or gradients from clients and redistributing the global model to clients for further learning. However, statistical heterogeneity across clients and partial client participation during training pose challenges to the learning process in FL. Furthermore, sparse training and inference are desirable to improve generalizability \citep{tibshirani1996regression} and enhance computational and communication efficiency in FL, thereby posing an additional challenge of discovering sparse models \citep{wang2021fieldguidefederatedoptimization}. 

A common approach to inducing sparsity relies on $L_1$ and $L_2$ norms\footnote{ For $\theta\in R^{d}$, $L_1$ norm is $\|\theta\|_1 = \sum_{i=1}^{d} |\theta_i|$ and $L_2$ norm is $\|\theta\|_2 = \left( \sum_{i=1}^{d} \theta_i^2 \right)^{1/2}$.}  for regularization, which depend directly on parameter magnitudes and offer varying levels of shrinkage. In contrast, using a magnitude-independent $L_0$ pseudo-norm is advantageous because it imposes a constant penalty on nonzero parameters and is useful for learning a model with a desired parameter density $\rho$. The Lagrangian for the $L_0$ density-constrained optimization problem in FL can be defined as: 
\begin{equation}\label{prb_def}
\mathfrak{L}(\theta,\lambda) =
\sum_{c=1}^{C} \frac{n_c}{N} \mathcal{L}^{(c)}(\theta)
+ \lambda \left(\|\theta\|_0 - \rho |\theta| \right), \quad
\|\theta\|_0 = \sum_{j=1}^{|\theta|} \mathbb{I}[\theta_j \neq 0],
\end{equation}
where, $\mathcal{L}^{(c)}(\theta)$ denotes the normalized loss at client $c$, defined as:
\begin{equation}
\mathcal{L}^{(c)}(\theta) =
\frac{1}{n_c} \sum_{i=1}^{n_c}
\ell\big(h(x_i^{(c)}; \theta), y_i^{(c)}\big).
\end{equation}
Here, $C$ is number of clients in FL, each holding a local dataset $(D^{(c)})_{c=1}^{C} = (X^{(c)}, Y^{(c)})$, where $X^{(c)} \in \mathbb{R}^{n_c \times in}$, $Y^{(c)} \in \mathbb{R}^{n_c\times out}$, and $\sum_{c=1}^{C} n_c = N$. We assume a model $h(x;\theta): \mathbb{R}^{in} \rightarrow \mathbb{R}^{out}$ and a loss function $\ell(h(x;\theta), y)$, where $x \in \mathbb{R}^{in}$, $y \in \mathbb{R}^{out}$, and $\theta \in \mathbb{R}^{d}$. The above formulation with Lagrange multiplier $\lambda $ leads to a min-max optimization problem that can, in principle, be solved using gradient descent-ascent. However, the non-differentiability of the $L_0$ pseudo-norm complicates optimization within standard gradient-based learning frameworks.

\citet{Louizos2017LearningSN} introduce a reparameterization using stochastic gates $z \in \mathbb{R}^{p}$ with $\theta = \tilde{\theta} \odot z$\footnote{$\odot$ denotes the elementwise product \citep{horn1990hadamard}.} as the effective parameters of the model. By assuming a hard Concrete distribution over $z$ (an approximation of the Bernoulli distribution), the $L_0$ norm is approximated by the expected number of active gates, i.e., $\sum_{j} \mathbb{E}_{q(z)}[z_j]$, enabling gradient-based optimization of the $ L_0$-regularized objective. \citet{GallegoPosada2022ControlledSV} extended this framework by using an $L_0$ density constraint and solving the resulting min--max problem, thereby enabling the user to define the target sparsity in a centralized setting. \citet{huthasana2025federatedlearningl0constraint} further adapted this approach to sparsity learning in the FL context, accounting for heterogeneity in data and client participation. 

However, learning sparse models is challenging, particularly in the small-sample high-dimensional regime $(d \gg N)$ \citep{ent1,ent2,ent3,ent4}, where the optimization is unstable, resulting in multiple solutions of competing parameter configurations with poor sparsity recovery and generalization behaviour to unseen test data. \citet{bao2022fast} work on statistical recovery in a low sample--high dimensional regime, but a low sample size for the individual client relative to the number of parameters is considered, while the total sample size N far exceeds the number of parameters. We aim to study a low total sample size $N$ relative to the number of parameters $d=|\theta|$ similar to a centralized setting, which is extremely challenging under data and client participation heterogeneity in FL. 

\citet{Louizos2017LearningSN} points to penalizing relative entropy or the Kullback-Leibler divergence $\mathrm{KL}(q(z) \| p(z))$ in the optimization where $p(z)$ is the prior and $q(z)$ is the approximate posterior for uncertainty exploration. The concept of entropy regularization is not new, and it is commonly used to encourage diversity and exploration at the decision level, distributions of class or action, in reinforcement learning (RL), and adapted to uncertainty exploration in input space, distributions of latent variables dependent on data, in variational auto--encoders \citep{ent5,ent6}. It is also used in Bayesian inference in centralized and FL settings \citep{ent7,ent8}. In their approach, the hard concrete gate distributions $q(z)$ are not directly dependent on the data and can be sampled independently of it post-training. In this work, the formulations at \citet{Louizos2017LearningSN,GallegoPosada2022ControlledSV} are adapted with relative-entropy penalization in a communication-efficient FL setting to explore uncertainty in non-zero parameter configurations and show that it consistently outperforms iterative hard-thresholding-based pruning during training \citep{tong_federated_2022} and post-training pruning of a dense model using the classic federated averaging algorithm \citep{1}.    

The remainder of this paper is organized as follows. We first present the entropy-regularized $L_0$-constrained formulation for federated optimization, then the proposed distributed algorithm, followed by experiments in heterogeneous FL settings, and conclude. 

\section{Formulation}
Assuming a model $h(x;\theta): \mathbb{R}^{in} \rightarrow \mathbb{R}^{out}$ and a loss function $\ell(h(x;\theta), y)$, where $x \in \mathbb{R}^{in}$, $y \in \mathbb{R}^{out}$, and $\theta \in \mathbb{R}^{d}$, consider a centralized dataset $D = (X, Y)$ with $X \in \mathbb{R}^{N \times in}$ and $Y \in \mathbb{R}^{N\times out}$. Using \citep{Louizos2017LearningSN, GallegoPosada2022ControlledSV}, the min--max objective can be defined using the expectation of the loss with respect to the distribution of gates, and the $L_0$ pseudo-norm approximated by the expected number of active gates, i.e., $\sum_j \mathbb{E}_{q(z \mid \phi)}[z_j]$ as shown in eq. \ref{lag_den_con_1}. Since each $z_j$ is a deterministic transformation of parameter-free noise, the expectation can be optimized using Monte Carlo sampling and reparameterized gradients \citep[ch.~3.3.3]{ranganath2017black}.

\begin{equation}\label{lag_den_con_1}
\hat{\mathfrak{L}}(\tilde{\theta}, \phi, \lambda) =
\frac{1}{R} \sum_{r=1}^{R}
\left[
\frac{1}{N} \sum_{i=1}^{N}
\ell\big(h(x_i; \tilde{\theta} \odot z^{(r)}), y_i\big)
\right]
+ \lambda \left( \sum_{j=1}^{|\theta|}
\mathbb{E}_{q(z \mid \phi)}[z_j] - \rho |\theta|
\right).
\end{equation}

In a federated learning (FL) setting with $C$ clients holding datasets $(D^{(c)})_{c=1}^{C} = (X^{(c)}, Y^{(c)})$, we consider a reparameterized linear model $h(x;\tilde{\theta} \odot z): \mathbb{R}^{in} \rightarrow \mathbb{R}^{out}$ and a loss function $\ell(h(x;\tilde{\theta} \odot z), y)$. Here, $X^{(c)} \in \mathbb{R}^{n_c \times in}$, $Y^{(c)} \in \mathbb{R}^{n_c\times out}$, $\sum_{c=1}^{C} n_c = N$, $x \in \mathbb{R}^{in}$, $y \in \mathbb{R}^{out}$, and $\theta = \tilde{\theta} \odot z \in \mathbb{R}^{in}$. The gate parameters are defined as $\phi = \log \alpha \in \mathbb{R}^{in}$.

The Lagrangian corresponding to the entropy regularized $L_0$ density-constrained optimization problem is:
\begin{equation}\label{prb_def2.3}
\hat{\mathfrak{L}}(\tilde{\theta}, \phi, \lambda) =
\sum_{c=1}^{C} \frac{n_c}{N} \mathcal{L}^{(c)}(\tilde{\theta}, \phi)
+ \lambda \left(
 \sum_{j=1}^{|\theta|}
\mathbb{E}_{q(z \mid \phi)}[z_j] - \rho|\theta|
\right)+T\sum_{j=1}^{|\theta|}KL(q(z_j|\phi)||p(z_j|\phi_{init})).
\end{equation}
Here, $p(z|\phi_{init})$ is a prior and is also a hard concrete distribution, and  $\mathcal{L}^{(c)}(\tilde{\theta}, \phi)$ denotes the Monte Carlo estimate of the normalized loss at client $c$, defined as:
\begin{equation}
\mathcal{L}^{(c)}(\tilde{\theta}, \phi) =
\frac{1}{R} \sum_{r=1}^{R}
\frac{1}{n_c} \sum_{i=1}^{n_c}
\ell\big(h(x_i^{(c)}; \tilde{\theta} \odot z^{(r)}), y_i^{(c)}\big).
\end{equation}
The stochastic gates $z$ are sampled using the Hard Concrete distribution, applying a hard-sigmoid transformation to a stretched Binary Concrete random variable \citep{Louizos2017LearningSN, Maddison2016TheCD}, defined as:
\begin{align}\label{eq4mainp}
s &= \sigma\left(
\frac{\log\frac{u}{1-u} + \log \alpha}{\beta'}
\right), \quad u \sim \mathcal{U}(0, 1),\notag \\
\bar{s} &= s(\zeta - \gamma) + \gamma, \quad
z = \min(1, \max(0, \bar{s})).
\end{align}
The expectation of a gate being active \footnote{$\mathbb{E}_{q(z \mid \phi)}[z_j] =
1 - Q(\bar{s}_j \leq 0 \mid \phi_j) =
\sigma\left(
\log \alpha_j - \beta' \log\left(-\frac{\gamma}{\zeta}\right)
\right)$} is derived at \citet{Louizos2017LearningSN} using the cumulative distribution function $Q(\bar{s})$. We introduced $T\geq 0$, which we treat as a constant with or without decay to penalize entropy. The resulting min--max optimization problem is:
\begin{equation}\label{centralFL}
\tilde{\theta}^*, \phi^*, \lambda^* =
\arg \min_{\tilde{\theta}, \phi}
\arg \max_{\lambda \geq 0}
\hat{\mathfrak{L}}(\tilde{\theta}, \phi, \lambda).
\end{equation}
The parameters $\tilde{\theta}$ and $\phi = \log \alpha$ are jointly optimized using gradient descent with reparameterized gradients. $\lambda$ is updated via gradient ascent and a restart strategy of resetting its value to 0 as and when the sparsity constraint is satisfied \citep{GallegoPosada2022ControlledSV}. Since the hard concrete distribution is a continuos approximation of Bernoulli distribution the computation of $KL(q(z_j)||p(z_j))$ involves an additional term involving $\bar{s} \text{, and thus, }z\in(0,1)$ which can be computed using truncated distribution $q(\bar{s}|\bar{s}\in (0,1))$ or a Monte-carlo estimate of the same. We used the closed-form expressions presented in the appendix~\ref{C} provided at \citet{Louizos2017LearningSN}. 

\section{Algorithm}
We use the notations $\mathcal{L}_{\text{Con}}(\phi)$ and $\mathcal{L}_{\text{KL}}(\phi)$ for $L_0$ density constraint and the $KL(q(z)||p(z))$ in \eqref{prb_def2.3}~\citep{huthasana2025federatedlearningl0constraint}. The Lagrangian can then be written as:
\begin{equation}\label{cent_adapt}
\hat{\mathfrak{L}}(\tilde{\theta}, \phi, \lambda) =
\sum_{c=1}^{C} \frac{n_c}{N} \mathcal{L}^{(c)}(\tilde{\theta}, \phi)
+ \lambda \, \mathcal{L}_{\text{Con}}(\phi)+T \, \mathcal{L}_{\text{KL}}(\phi).
\end{equation}
\citet{1} propose federated averaging (\texttt{FedAvg}), a distributed algorithm for learning a global model via synchronous updates from clients using gradient averaging. In this setting, a central server coordinates training across $C$ clients each holding a local dataset $D^{(c)}$. Each client performs stochastic gradient descent (SGD) updates locally for few iterations before communicating the parameters to the server, to reduce the communication between server and clients by increasing computation at clients. The server aggregates the parameters by averaging to obtain the global model.

At iteration $t$, each client performs the following updates:
\begin{align}
\tilde{\theta}_{c}^{t+1} &= \tilde{\theta}_{c}^{t}
- \eta_{\tilde{\theta}}
\nabla_{\tilde{\theta}} \mathcal{L}^{(c)}(\tilde{\theta}^{t}, \phi^{t}), \\
\phi_{c}^{t+1} &= \phi_{c}^{t}
- \eta_{\phi} \left(
\nabla_{\phi} \mathcal{L}^{(c)}(\tilde{\theta}^{t}, \phi^{t})
+ \lambda \text{ }\nabla_{\phi} \mathcal{L}_{\text{Con}}^{(c)}(\phi^{t})+T\text{ }\nabla_{\phi} \mathcal{L}_{\text{KL}}^{(c)}(\phi^{t})
\right), \\
\lambda_{c}^{t+1} &= \lambda_{c}^{t}
+ \eta_{\lambda} \, \mathcal{L}_{\text{Con}}^{(c)}(\phi^{t}).
\end{align}
At each round, a fraction $\gamma_c$ of clients is selected uniformly at random, resulting in $K = \lfloor \gamma_c C \rfloor$ participating clients. In practice, a few mini-batches of uniform size $B$ sampled iteratively at each client to locally update the model $n(B)$ times per communication round or a full pass over client data amount to a local epoch. The server gathers all updates from clients and performs a synchronous update of global model , using the averages of the updates from clients, after each communication round or global epoch keeping $T$ constant or decaying it according to a predetermined schedule. The Lagrange parameter $\lambda$ is reset to zero when the constraint is satisfied \citep{GallegoPosada2022ControlledSV}. The aggregation weights $w_k$ for clients sampled in an epoch or round can be uniform or proportional to the number of samples each client holds.

This approach enables learning a global sparse model in FL with entropy regularization and an $L_0$ constraint using probabilistic gates. We refer to this variant of FedAvg as \texttt{E-FLoPS}, where \texttt{E} stands for entropy regularization. 
The learning rates $\eta_{\tilde{\theta}}$ and $\eta_{\phi}$ needs to be appropriately tuned. The learning rate $\eta_{\lambda}$ for the Lagrange parameter updates is set in the order of $1/|\theta|$. The temperature $T\in (0,\infty)$ is also initialized at $1/|\theta|$ and increased if needed to encourage uncertainty exploration as the total sample size N decreases.   The gate parameters $\phi = \log \alpha$ are initialized from a normal distribution with mean $\log \rho_{\text{init}} - \log(1 - \rho_{\text{init}})$ and variance $0.01$, where $\rho_{\text{init}}$ controls the initial density. The target density is denoted by $\rho_{\text{targ}} = \rho$. The Hard Concrete distribution parameters are set to $\gamma = -0.1$, $\zeta = 1.1$, and $\beta'$ is recommended to be set at $0.66$, following \citet{Louizos2017LearningSN}. The Lagrange multiplier is initialized as $\lambda = 0$. Client participation variability is simulated by randomly selecting a fraction of clients at each training round.

\texttt{E-FLoPS} achieves test-time sparsity by using deterministic gates $\hat{z}$ sampled without noise or smoothing \citep{GallegoPosada2022ControlledSV}. For exact sparsity, the raw parameters $\tilde{\theta}$ for the top-$m$ indices of effective parameters $\theta$ are pruned, where $m = \lfloor \rho_{\text{targ}} \cdot |\theta| \rfloor$. This pruning mechanism can be applied from a pre-defined threshold of epochs, referred to as the \emph{prune start} epoch, which is set to $0$ for \texttt{E-FLoPS} to improve communication efficiency via serialization for message compression \citep{huthasana2025federatedlearningl0constraint}. At this stage, the serialized mean gate values, from repeated sampling, corresponding to top-$m$ indices of gates $z$ are retained, and the rest of the gates are replaced with their average, denoted by $z^{\text{avg}}_{-m}$. Only the pruned $\tilde\theta$ and $z$, along with their non-zero indices and the scalar $z^{\text{avg}}_{-m}$, are communicated. Upon reception, parameters are reconstructed via $\theta = \tilde{\theta} \odot z$ and aggregated, and $\phi$ is recovered from $z$ using $\scriptstyle\phi = \beta' \log\left(\frac{z}{1 - z}\right)$, ignoring the noise component of the gates. The communication cost reduction by this approach in FL is significant for small $\rho_{\text{targ}}$ with minimal meta data overhead.

\begin{algorithm}
\caption{FedAvg variant \texttt{E-FLoPS}. $E$ and $B$ denote the number of epochs and the mini-batch size.}
\label{algo:flops-avg}
\begin{algorithmic}[1]

\State \textbf{Initialization:} $(\tilde{\theta}^{(0)}, \phi^{(0)})$, compute $\theta^{(0)} = \tilde{\theta}^{(0)} \odot z^{(0)}$

\For{epoch $b = 1$ to $E$}
    \State Sample client subset $S_t$

    \For{\textbf{each} client $k \in S_t$}
        \State $(k,\tilde\theta_k, z_k,\lambda_{k}) \leftarrow$ \textbf{ClientCompute}$(k, \tilde\theta^{(t)}, z^{(t)},\lambda^{(t)})$
    \EndFor
    
    \State \textbf{Server aggregation:}
    \begin{align*}
    \theta &= \sum_{k \in S_t} w_k \tilde\theta_k \\
    z &= \sum_{k \in S_t} w_k z_k
    \end{align*}

    \State Reconstruct $\tilde{\theta} = \theta \oslash z$\footnotemark and recover $\phi$ from $z$ using $\phi = \beta' \log\left(\frac{z}{1 - z}\right)$

    \State Perform server-side updates on $(\tilde{\theta}, \phi,\lambda)$
    \State Either decay or keep $T$ constant

    \If{$b >$ prune start}
        \State prune $\tilde\theta$ using top-$m$ $\theta$ top-$m$ indices of $\theta$
        \State prune $z$ with tail statistic $\scriptstyle z^{\text{avg}}_{-m}$
        \State Serialize and communicate
    \EndIf

\EndFor

\State \textbf{ClientCompute}$(k, \tilde\theta, z,\lambda)$:
\State \quad Recover $\phi_k$ from $z$ using $\scriptstyle\phi = \beta' \log\left(\frac{z}{1 - z}\right)$ 
\State \quad Sample mini-batches $b \sim D^{(k)}$ of size $B$
\State \quad Perform $n(B)$ local SGD steps on $(\tilde{\theta}_k, \phi_k)$ and ascent for $\lambda_{k}$
\State \quad \Return Communicate pruned and serialized $(\tilde{\theta}_k, \phi_k)$ and $\lambda_k$
\end{algorithmic}
\end{algorithm}
\footnotetext{$\oslash$ represents an element wise division.}
\section{Experiments}

We include experiments on synthetic and real-world datasets. We evaluate true sparsity recovery in linear regression (LR) on synthetically generated data, convolutional neural network (CNN) on the MNIST digit classification data, and softmax multi-class classifier on the Golub leukemia cancer classification data. We compare our method with the federated iterative hard thresholding algorithm (\texttt{Fed-IHT}) proposed by \citet{tong_federated_2022}, where a hard-thresholding operation is used to retain only top-$m$ parameters in absolute magnitude at each iteration or epoch along with federated averaging to enforce sparsity, and with classic federated averaging (\texttt{Fed-Avg}) with dense training and pruning after the last training epoch. For \texttt{FedAvg} the training time performance on test data is evaluated using the top-$m$ parameters after each epoch though the training is dense without imposing sparsity. An approximate upper bound is established by centralized training by pooling data from clients with distributional shifts. A tuning phase, marked by a vertical dotted line in all figures in the experiments section, is conducted to further improve the statistical performance of the models with the fixed sparse support discovered by the end of the training phase.  

In FL, comunication is a bottle neck and needs to be minimized. The federated averaging variant \texttt{E-FLoPS} enables sparse communication throughout training without compromising statistical performance. The theoritical uplink and downlink communication costs can be estimated as multiples of the message size and the number of communication rounds, assuming 4 bytes per parameter and index each. The total two way communication cost per client in each server round in FL is,
\[
\resizebox{0.9\textwidth}{!}{$
\texttt{FedAvg}: \text{epochs} \times 4 |\theta|,
\text{ }
\texttt{E-FLoPS}: \text{epochs} \times 4 \cdot (2\rho_{\text{targ}} |\theta|),
\text{ and }
\texttt{FedIter-HT}: \text{epochs} \times 4 \cdot (\rho_{\text{targ}} |\theta|)
$}.
\]
Thus, the communication cost of \texttt{E-FLoPS} is more than \texttt{FedIter-HT}, but significantly lower than dense training via \texttt{FedAvg}. For large model sizes $|\theta|$ and small target densities $\rho_{\text{targ}}$, the gap between \texttt{FedAvg} and \texttt{E-FLoPS} becomes substantial, while small with that of \texttt{FedIter-HT}.

The experiments are conducted on an Apple MacBook with an M4 Pro chip (12-core CPU) and 24\,GB unified memory, running macOS 15.5. The implementation uses Flower (v1.29.0) framework for simulation of FL and PyTorch (v2.7.0) framework for training, with Python (v3.12.7).

\subsection{Experiments on Synthetic Data}

We generated synthetic data for sparse linear regression following the procedure from \citet{bertsimas_sparse_2020} for a range of $\scriptstyle \frac{N}{|\theta|}$. For a parameter vector dimension $d=|\theta|$, each row $x_i \in \mathbb{R}^{d}$ of $X \in \mathbb{R}^{N \times d}$ is sampled from a zero-mean Gaussian distribution with covariance matrix $\Sigma$. We use a Toeplitz covariance structure defined as $\Sigma_{ij} = \rho_{\text{cor}}^{|i-j|}, \quad i,j = 1, \ldots, d$. An $m$-sparse coefficient vector $w_{\text{true}} \in \mathbb{R}^{d}$ is constructed, where $m = \lfloor \rho \cdot d \rfloor$. A subset of indices $S_m \subseteq \{1, \ldots, d\}$ is selected uniformly at random, and coefficients are assigned as
$(w_{\text{true}})_j \sim \mathrm{Unif}\{-1,1\} \text{ for } j \in S_m \text{ and } (w_{\text{true}})_j = 0 \text{ otherwise}$. The response vector is generated as $y = X w_{\text{true}} + \varepsilon$, where $\varepsilon \sim \mathcal{N}(0, \sigma^2 I_N)$. The signal-to-noise ratio (SNR) is defined as $\scriptstyle\mathrm{SNR} = \frac{\|X w_{\text{true}}\|_2^2}{\|\varepsilon\|_2^2}$, and the noise level is set to $\scriptstyle\sigma = \frac{\|X w_{\text{true}}\|_2}{\sqrt{\mathrm{SNR}} \, \sqrt{N}}$. For $\scriptstyle|\theta|=1000$, we generate a number of total training samples N for varying ratios of $\scriptstyle\frac{N}{|\theta|}$ with $\scriptstyle5000$ samples of test data in all cases. The generated training data is then distributed among clients.

We employed affine shifting for attribute distribution skew, Dirichlet partitioning protocol (DPP) for skew in the number of samples available at a client \citep{reisizadeh2020robust,solans2024non}. A fraction of clients ($0.6$) is randomly sampled in each epoch to introduce participation heterogeneity \citep{reisizadeh2020robust}. We evaluate performance using mean squared error (MSE) and $R^2$. Experiments are conducted at a correlation level of $\rho_{\text{cor}}=0.2$ and an SNR of 20, with a true sparsity level of $95\%$ or target density of $5\%$.

\begin{figure}
\centering

\begin{minipage}{0.48\linewidth}
\centering
\includegraphics[width=\linewidth]{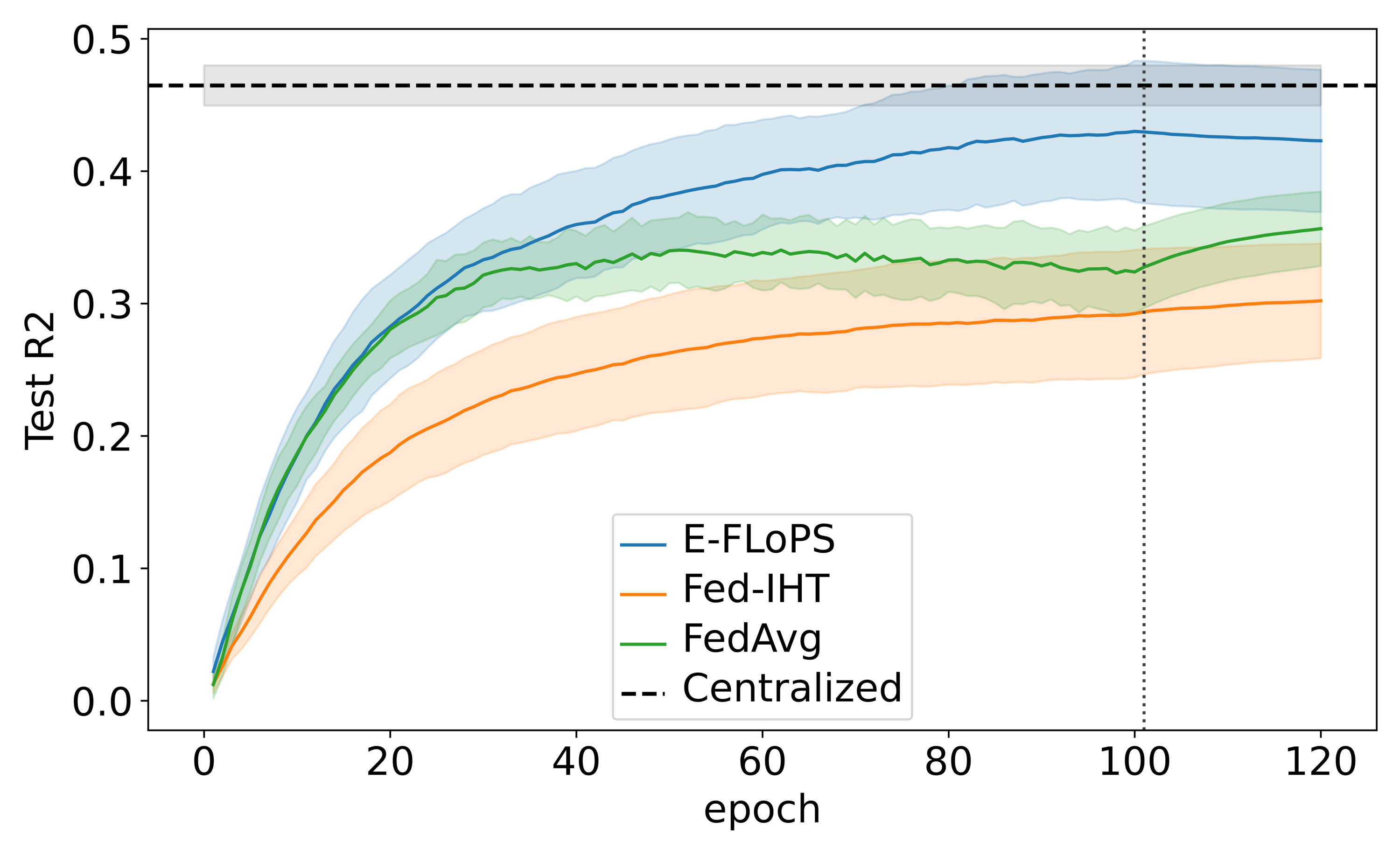}
\end{minipage}
\hfill
\begin{minipage}{0.48\linewidth}
\centering
\includegraphics[width=\linewidth]{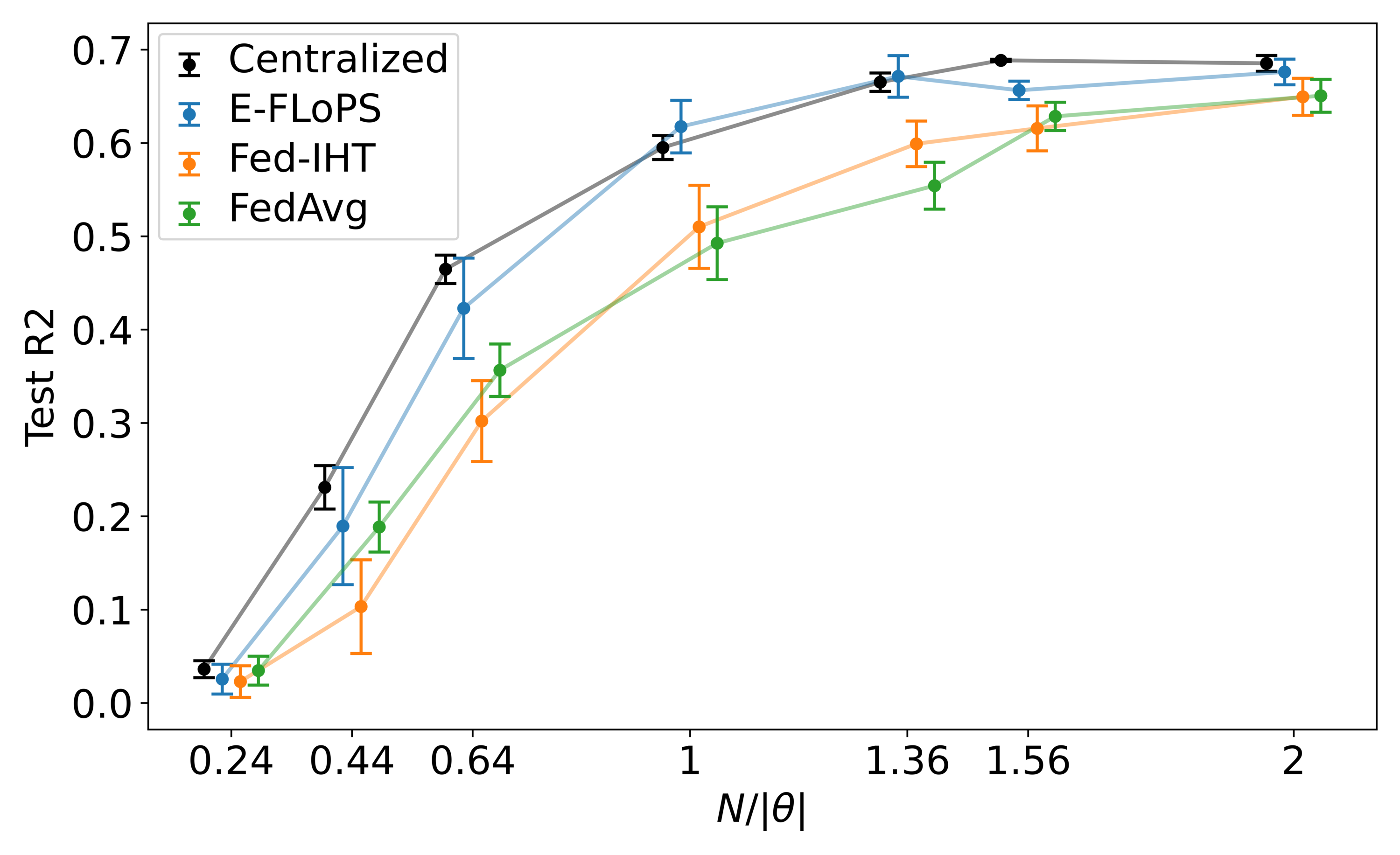}
\end{minipage}
\begin{minipage}{0.48\linewidth}
\centering
\includegraphics[width=\linewidth]{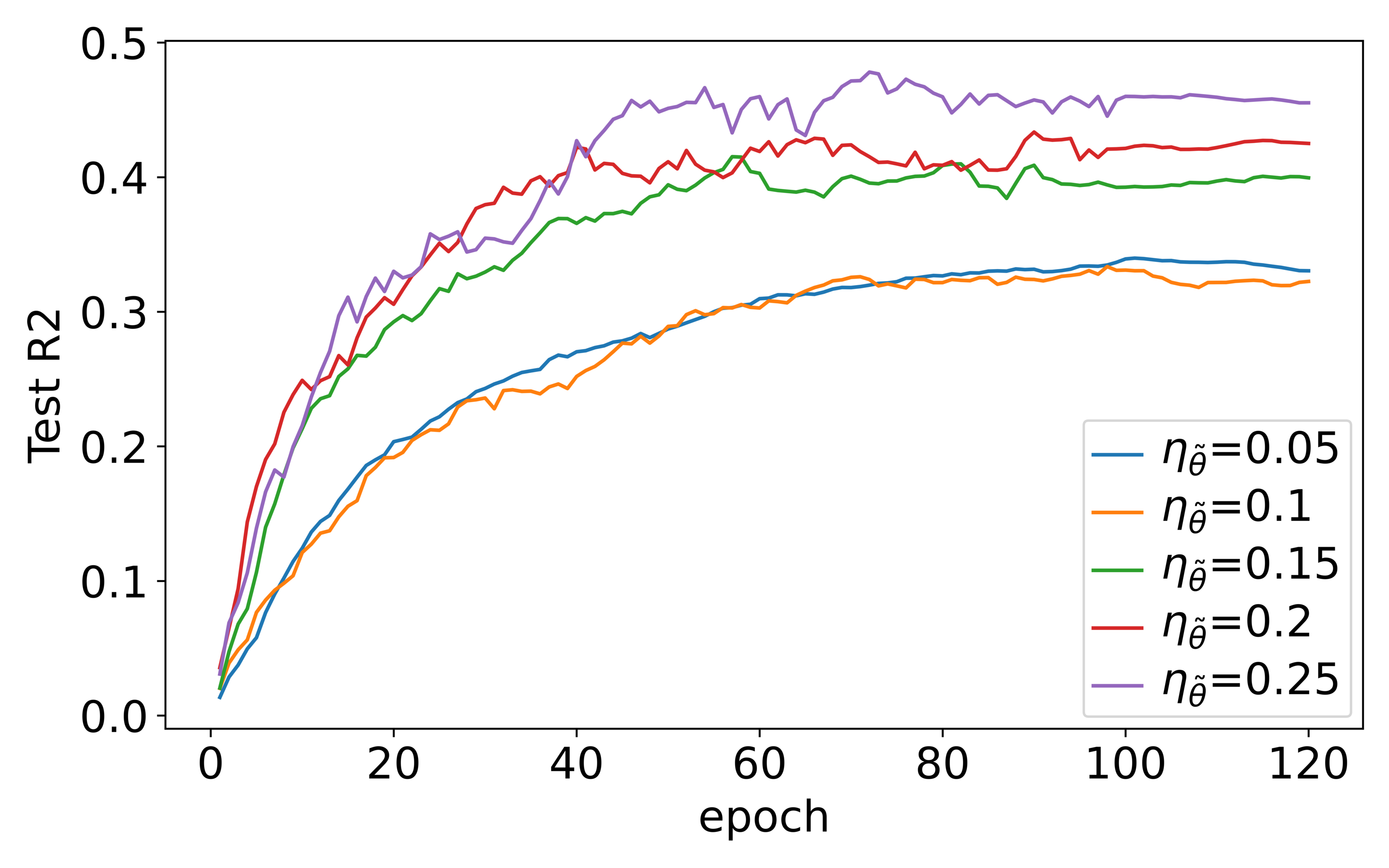}
\end{minipage}
\hfill
\begin{minipage}{0.48\linewidth}
\centering
\includegraphics[width=\linewidth]{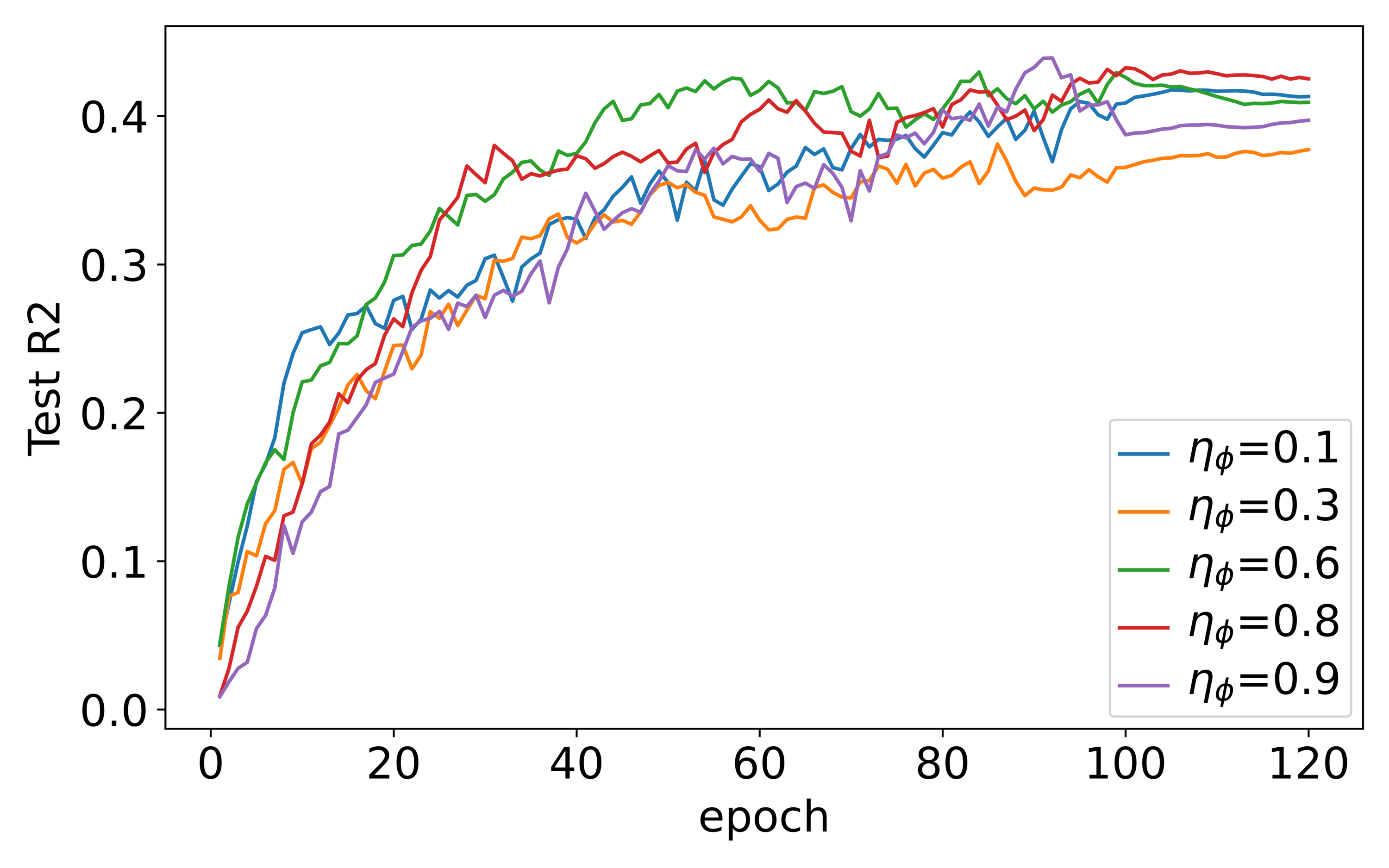}
\end{minipage}

\caption{The figures show (a) mean test $R^2$ and standard deviation for $\scriptstyle\frac{N}{|\theta|}=0.64$ over 30 runs of all the algorithms, (b) test $R^2$ over varying $\scriptstyle\frac{N}{|\theta|}$ by changing total number of samples available across all clients,(c) test $R^2$ at $\eta_{\phi}=0.85$ for varying $\eta_{\tilde{\theta}}$, and (d) test $R^2$ at $\eta_{\tilde{\theta}}=0.25$ for varying $\eta_{\phi}$.}
\label{syn_new}

\end{figure}

As shown in Figure~\ref{syn_new}, \texttt{E-FLoPS} achieves a higher $R^2$ on unseen test data than \texttt{FedIter-HT} at all $\scriptstyle\frac{N}{|\theta|}$. \texttt{E-FLoPS} also exceeds top-$m$ model of densely training \texttt{FedAvg} during training and the pruned and fine-tuned model discovered post training. The Figure~\ref{syn_new} also shows the stability of across varying learning rates of $\tilde{\theta}$, the raw parameters, and $\phi$, the parameters of the gates. For stable training of \texttt{E-FLoPS}, we fix $\scriptstyle\eta_{\lambda} = \frac{0.01}{|\theta|} $, and tune learning rates within the ranges $\eta_{\tilde{\theta}} \in [0.05, 0.5]$ and $\eta_{\phi} \in [0.1, 0.9]$.
 Table ~\ref{tab:support_recovery_scaling} shows the accuracy of sparsity recovery $\scriptstyle A(\theta)=\frac{TP}{(TP+FN)}$, the fraction of true features recovered. \texttt{E-FLoPS} has higher accuracy than both \texttt{Fed-IHT} and \texttt{FedAvg} except at smallest $\scriptstyle\frac{N}{|\theta|}$.  

\begin{table}[htbp]
  \caption{Final support recovery across total sample sizes ($N$) at $d=|\theta|$ showing mean $\pm$ standard deviation of $A(\theta)$ over 30 different run seeds. The federated method with the best support recovery is indicated in bold in each row, and the centralized method is shown as a reference upper bound.}
  \label{tab:support_recovery_scaling}
  \centering
  \scriptsize
  \resizebox{0.7\textwidth}{!}{%
  \begin{tabular}{ccccc}
    \toprule
    $\scriptstyle\mathbf{\frac{N}{d}}$ & \textbf{E-FLoPS} & \textbf{Fed-IHT} & \textbf{FedAvg} & \textbf{Centralized} \\
    \midrule
    0.24 & 0.140 $\pm$ 0.039 & 0.129 $\pm$ 0.039 & \textbf{0.179 $\pm$ 0.038} & 0.195 $\pm$ 0.027 \\
    0.44 & 0.379 $\pm$ 0.082 & 0.292 $\pm$ 0.044 & \textbf{0.427 $\pm$ 0.036} & 0.473 $\pm$ 0.027 \\
    0.64 & \textbf{0.626 $\pm$ 0.060} & 0.490 $\pm$ 0.055 & 0.566 $\pm$ 0.038 & 0.736 $\pm$ 0.026 \\
    1.00 & \textbf{0.828 $\pm$ 0.034} & 0.687 $\pm$ 0.072 & 0.741 $\pm$ 0.023 & 0.840 $\pm$ 0.016 \\
    1.36 & \textbf{0.925 $\pm$ 0.026} & 0.831 $\pm$ 0.032 & 0.803 $\pm$ 0.030 & 0.945 $\pm$ 0.010 \\
    1.56 & \textbf{0.920 $\pm$ 0.016} & 0.874 $\pm$ 0.032 & 0.890 $\pm$ 0.110 & 0.980 $\pm$ 0.000 \\
    2.00 & \textbf{0.941 $\pm$ 0.021} & 0.913 $\pm$ 0.028 & 0.917 $\pm$ 0.027 & 0.987 $\pm$ 0.010 \\
    \bottomrule
  \end{tabular}%
  }
\end{table}
\subsection{Experiments on Real Data}
\subsubsection{Image Classification}
We consider publicly available \emph{MNIST} data \citep{lecun1998mnist}: A multi-class classification dataset of handwritten digits with $28 \times 28$ gray-scale pixel values for each. We use a CNN with two \(5 \times 5\) convolutional layers of 6 and 16 channels, each followed by \(2 \times 2\) max pooling, and three fully connected layers with widths 120, 84, and 10 for this classification task resulting in a non-linear model with 44,426 parameters. The targeted density is $2.5\%$ ( sparsity of $97.5\%$ ).  

\begin{figure}
\centering
\resizebox{0.95\linewidth}{!}{%
\parbox{\linewidth}{%

\begin{minipage}{0.53\linewidth}
\centering
\includegraphics[width=\linewidth]{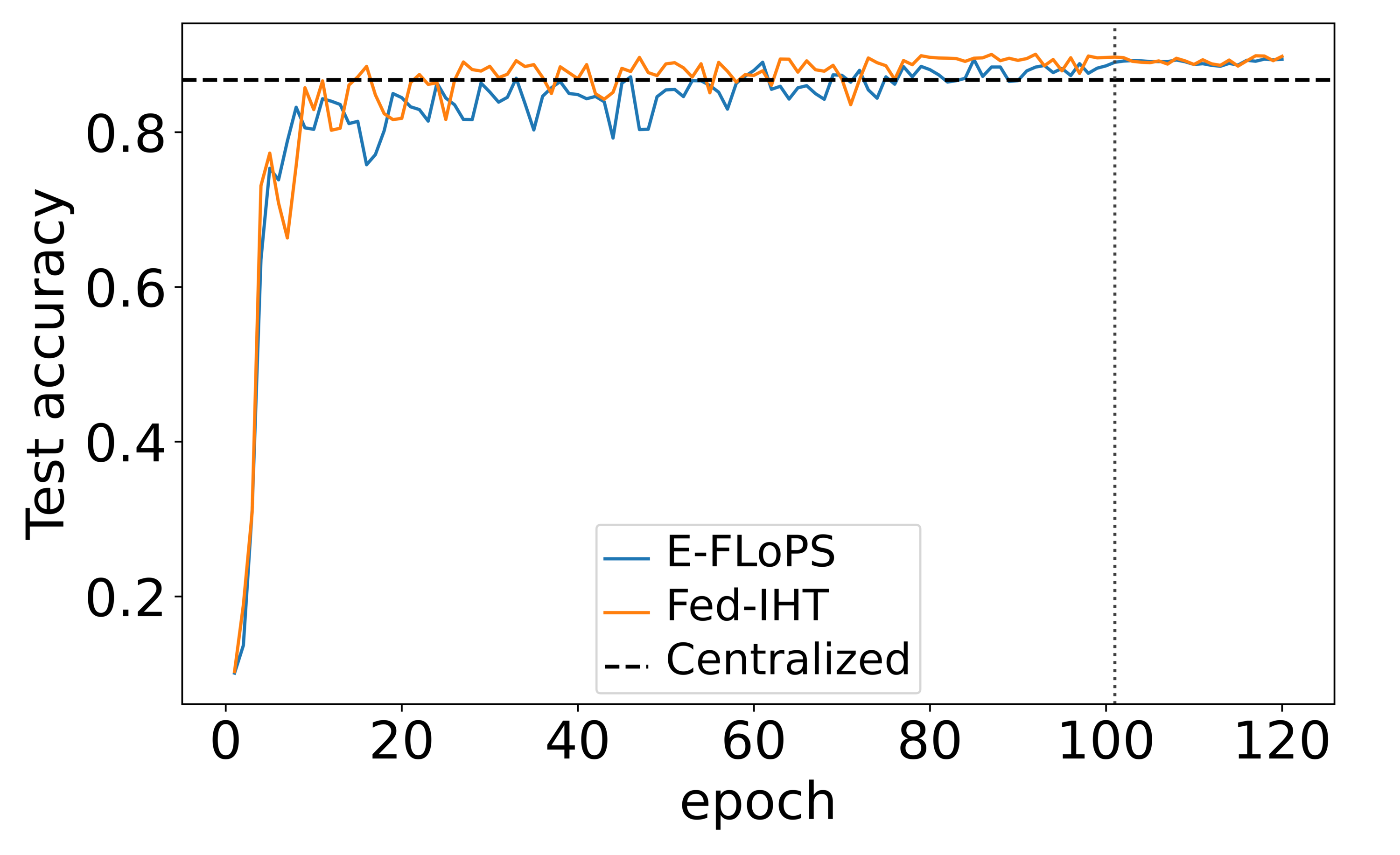}
\end{minipage}
\hfill
\begin{minipage}{0.46\linewidth}
\centering
\includegraphics[width=\linewidth]{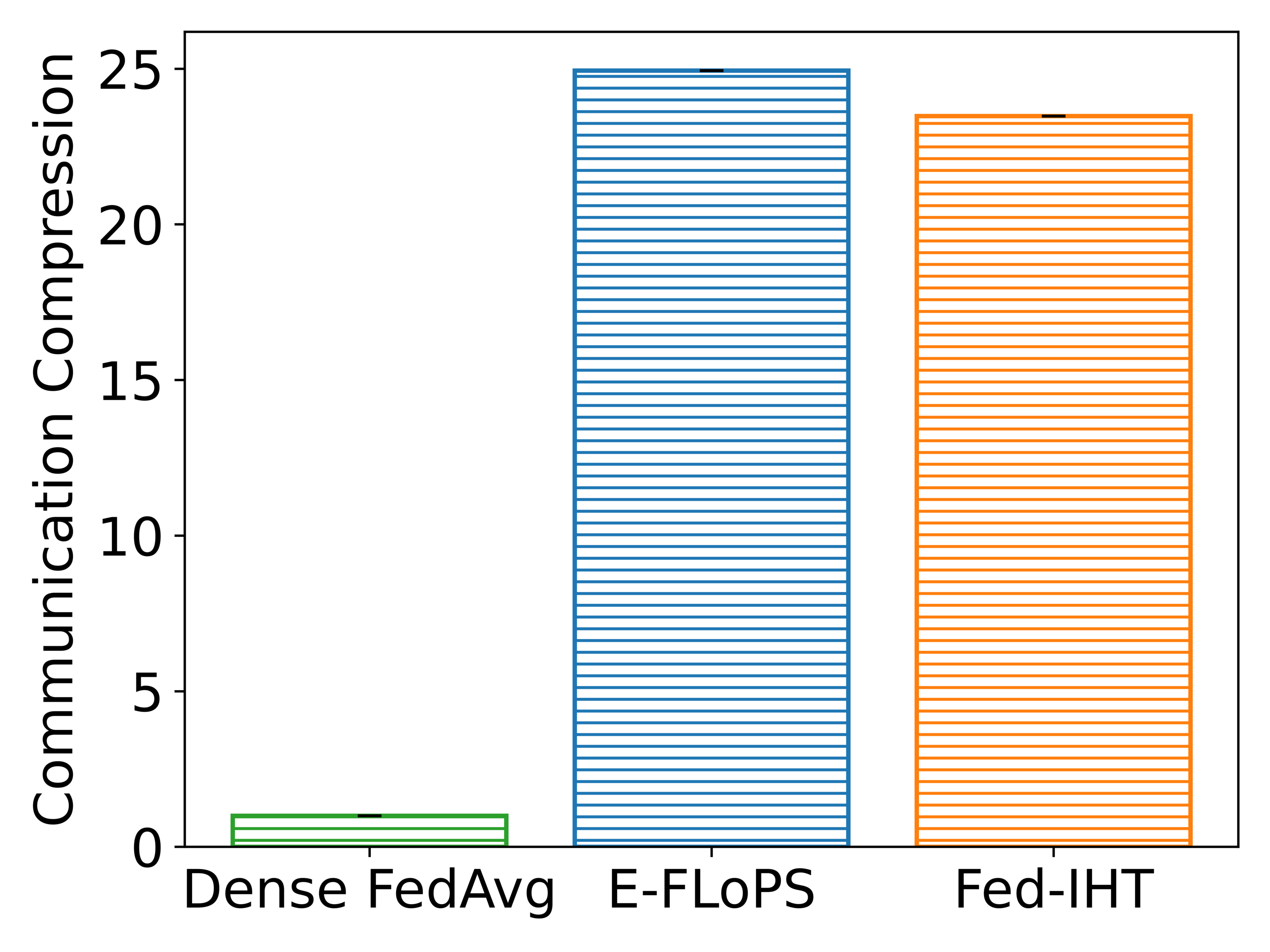}
\end{minipage}
\begin{minipage}{0.46\linewidth}
\centering
\includegraphics[width=\linewidth]{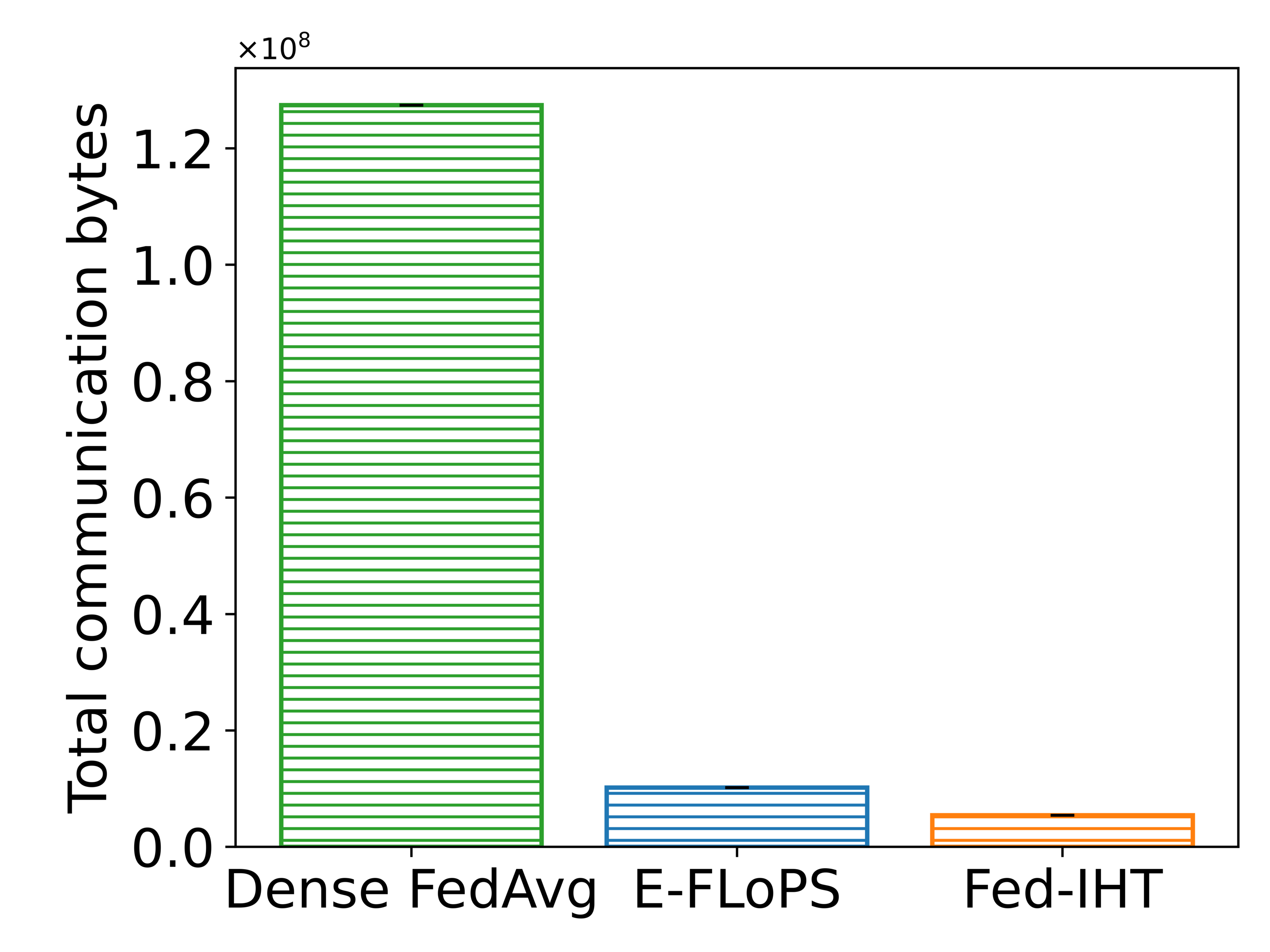}
\end{minipage}
\hfill
\begin{minipage}{0.46\linewidth}
\centering
\includegraphics[width=\linewidth]{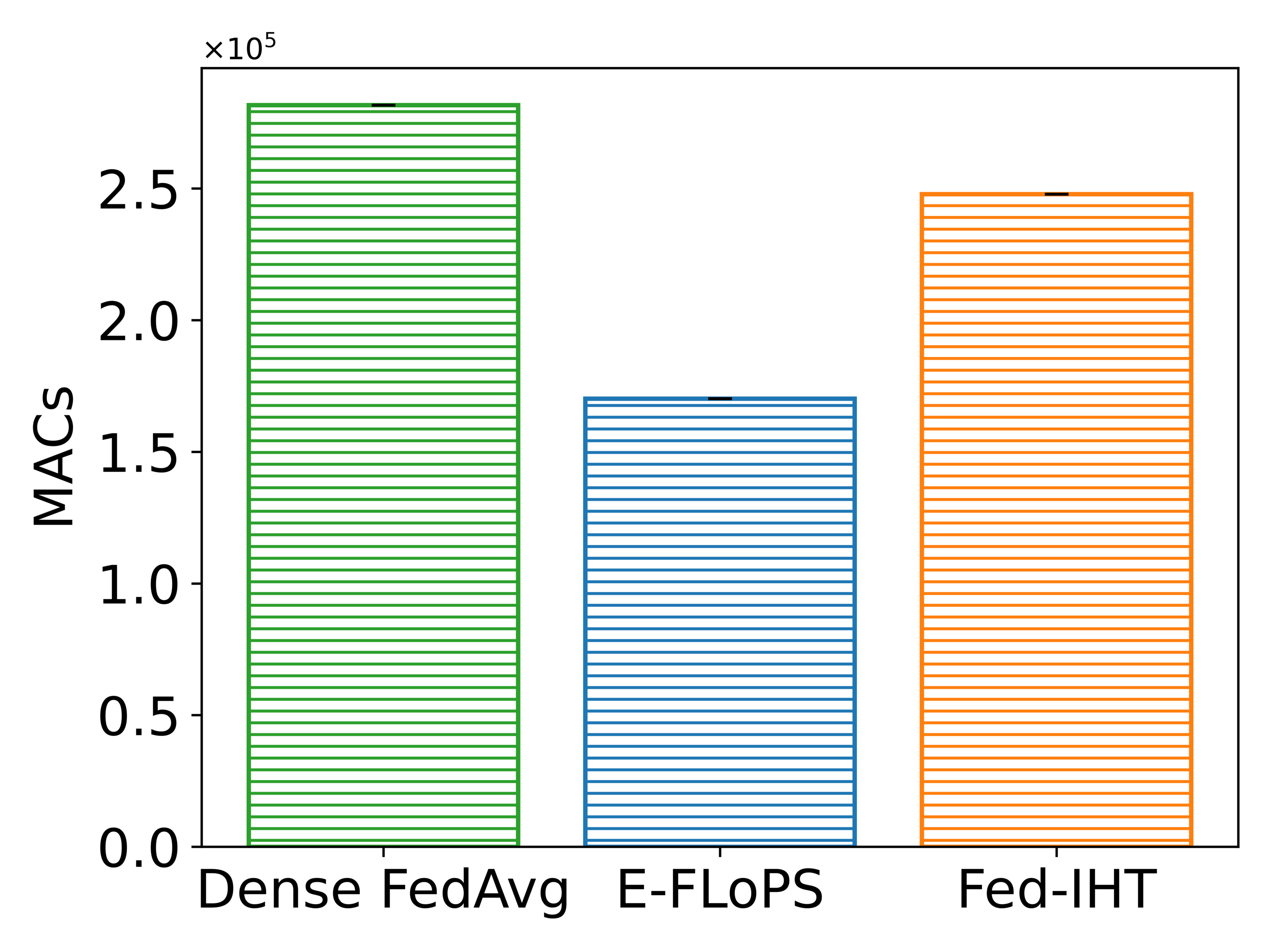}
\end{minipage}
}
}
\caption{The figures show: a) Test accuracy over epochs at $97.5\%$ sparsity at $\scriptstyle\frac{N}{\theta}\sim 0.02$, b) Communication compression compared to dense training, c) Total communication bytes between server and clients in training and fine-tuning, and d) MACs reduction.}
\label{real_tot}

\end{figure}

The data is downsampled per class at a ratio of $\sim$0.02, resulting in 1184 training samples and 10000 test samples. The data is distributed to clients using DPP for label skew and only a fraction of clients(0.6) are randomly sampled to train in each round. The affine shifts for further distributional heterogeneity are not performed in this case. For evaluation, we use cross-entropy (CE) loss with classification accuracy. Figure~\ref{real_tot} shows that \texttt{E-FLoPS} and \texttt{FedIter-HT} have similar test-time performance and communication efficiency. However, \texttt{E-FLoPS} shows a greater reduction of Multiply Accumulate Operations (MACs) leading to computational efficiency at inference time.   

\subsubsection{Leukemia Classification}
We use the publicly available Golub leukemia gene-expression dataset from the study on molecular classification of acute leukemia types, acute lymphoblastic leukemia (ALL) and acute myeloid leukemia (AML), using gene expression \citep{golub1999molecular}. The data consist of 72 patient samples with 3571 features after initial preprocessing. Each feature represents the measured expression level of a gene/probe in a patient sample; larger values indicate higher abundance of the corresponding gene transcript. These gene-expression profiles are used to classify samples as ALL or AML. This dataset combines a very small sample-to-dimension ratio, $\scriptstyle\frac{N}{|\theta|} = 0.02$, with dense, high-dimensional gene-expression features. It also represents a realistic setting in which data privacy may be a central concern, since the samples correspond to patient-level biomedical measurements.

\begin{figure}
\centering
\resizebox{0.95\linewidth}{!}{
\begin{minipage}{0.48\linewidth}
\centering
\includegraphics[width=\linewidth]{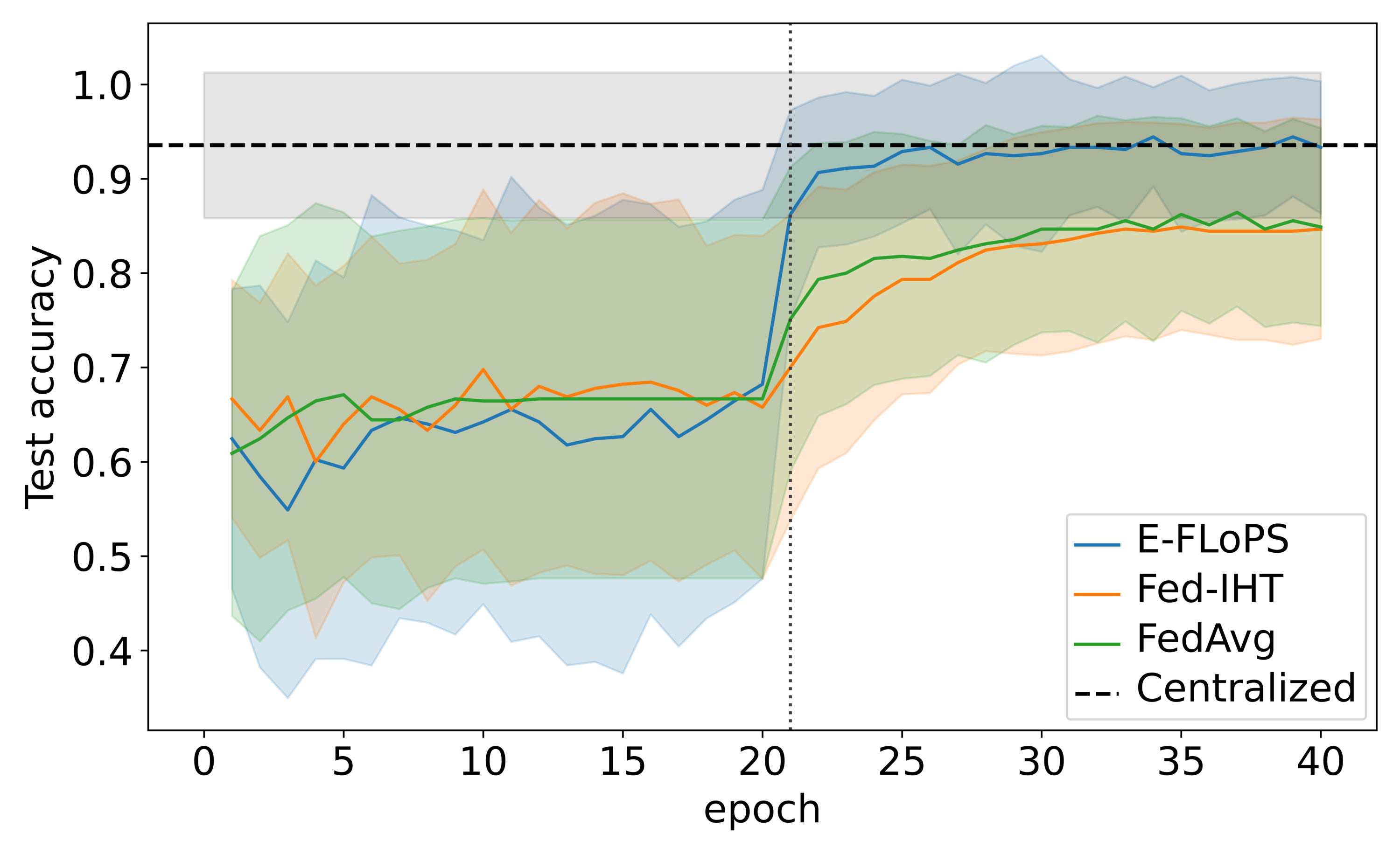}
\end{minipage}
\hfill
\begin{minipage}{0.48\linewidth}
\centering
\includegraphics[width=\linewidth]{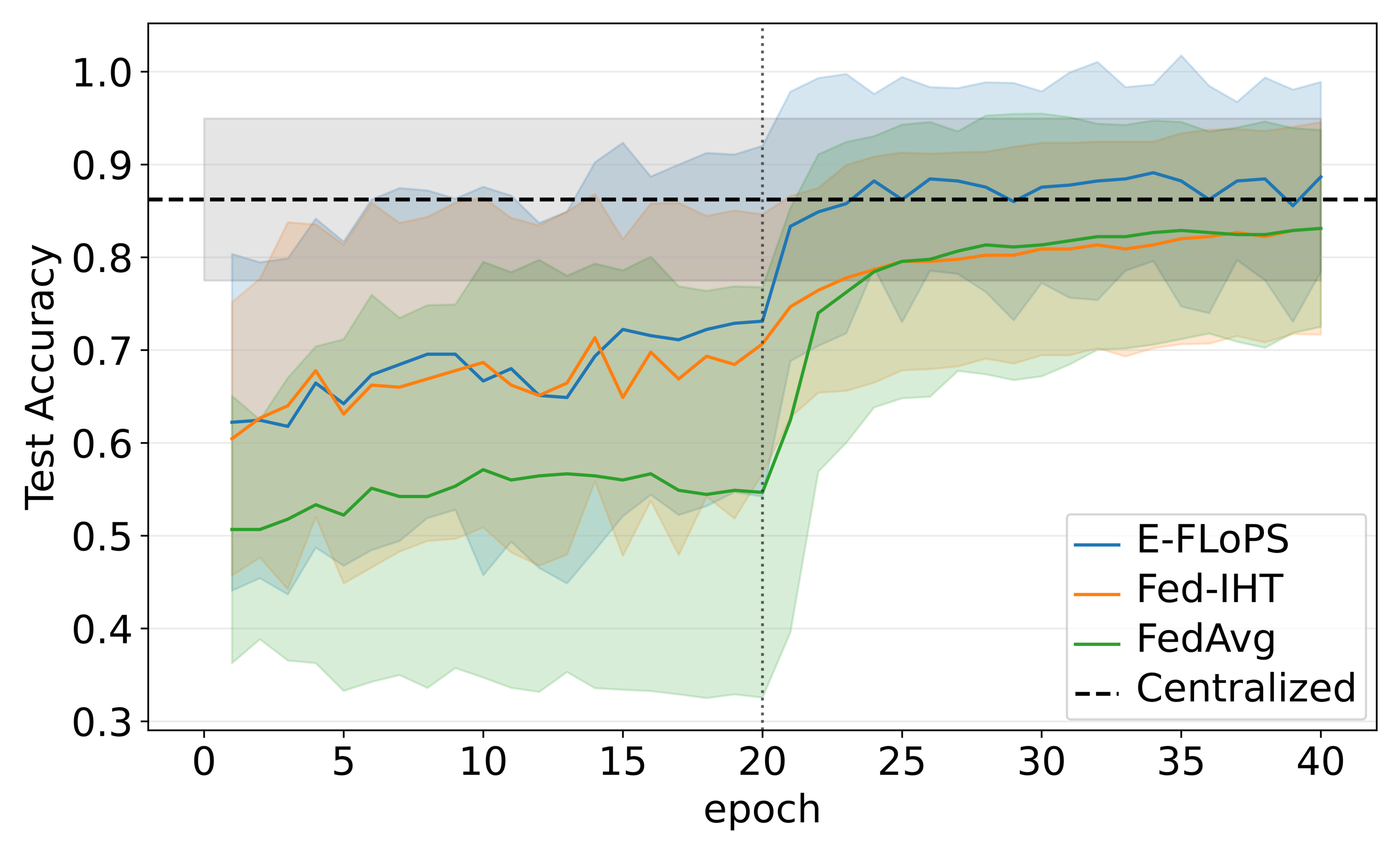}
\end{minipage}
}
\caption{The figure shows a) Test accuracy over epochs at $99.9\%$ sparsity at $\scriptstyle \frac{N}{|\theta|}\sim0.02$ at a fixed data/split seed and 30 different run seeds and b) Test accuracy over epochs at $99.9\%$ sparsity at $\scriptstyle\frac{N}{|\theta|}\sim0.02$ at a fixed run seed and 30 different data/split seeds. }
\label{real_tot2}
\end{figure}

We randomly select 15 samples for the test set and distribute the remaining samples across 5 clients using DPP at a fixed split seed. We train a softmax classifier at a target density of $0.1\%$ (99.9\% sparsity), with all clients participating. We do not synthetically alter client data using affine shifts. Figure~\ref{real_tot2} shows that \texttt{E-FLoPS} outperforms the other variants over 30 runs with different seeds for run-time stochastic processes. We confirmed that the difference between \texttt{E-FLoPS} and \texttt{Fed-IHT} is significant using a paired $t$-test, with a $t$-statistic of $3.79$ and a $p$-value of $7 \times 10^{-4}$. We also performed experiments across 30 train/test splits generated with different split seeds to ensure that the results were not driven by a particularly favorable random partition of the data. Across these splits, \texttt{E-FLoPS} again outperforms \texttt{Fed-IHT} and \texttt{FedAvg}. Table~\ref{tab:top_genes_by_method} presents the most frequently selected genes across different run seeds. The most frequently selected genes by \texttt{E-FLoPS}, cystatin C and myeloperoxidase, are supported by prior experimental research on their roles in promoting metastasis and cancer spread \citep{liang2013sparse}.

\begin{table}[htbp]
  \caption{The table shows the top genes selected by each method. Values in parentheses indicate the proportion of runs in which a particular gene is selected\protect\footnotemark.}
  \label{tab:top_genes_by_method}
  \centering
  \scriptsize
  \resizebox{\textwidth}{!}{%
  \begin{tabular}{llll}
    \toprule
    \textbf{E-FLoPS} & \textbf{Fed-IHT} & \textbf{FedAvg} & \textbf{Centralized} \\
    \midrule
    \textbf{cystatin C} (0.90) & ferritin, light polypeptide (0.70) & azurocidin 1 (0.33) & interleukin 8 (0.73) \\
    \textbf{myeloperoxidase} (0.80) & \textbf{cystatin C} (0.67) & interleukin 8 (0.33) & ferritin, light polypeptide (0.70) \\
    ferritin, light polypeptide (0.80) & azurocidin 1 (0.57) & ferritin, light polypeptide (0.30) & \textbf{myeloperoxidase} (0.67) \\
    interleukin 8 (0.53) & histocompatibility complex (0.57) & \textbf{myeloperoxidase} (0.27) & \textbf{cystatin C} (0.63) \\
    defensin, alpha 1 (0.43) & hemoglobin, beta (0.57) & \textbf{cystatin C} (0.23) & glycoprotein B (0.53) \\
    azurocidin 1 (0.33) & interleukin 8 (0.47) & interleukin 8 (0.17) & azurocidin 1 (0.47) \\
    histocompatibility complex (0.23) & \textbf{myeloperoxidase} (0.43) & elastase 2, neutrophil (0.17) & histocompatibility complex (0.37) \\
    \bottomrule
  \end{tabular}%
  }
\end{table}
\footnotetext{The gene "major histocompatibility complex, class I, B" is represented as "histocompatibility complex" for legibility in the Table~\ref{tab:top_genes_by_method}.}
\section{Conclusion}
Prior work has shown that sparse models can reduce communication overhead, memory footprint, and possibly inference cost, while also improving generalization. However, the challenge of sparse recovery in $(d \gg N)$ regime under heterogeneous FL conditions remains underexplored. We study entropy regularization of probabilistic gates for sparse FL under an $L_0$ density constraint. The proposed federated averaging algorithm, \texttt{E-FLoPS}, extends sparse federated optimization with a mechanism for uncertainty-driven exploration of sparse parameter configurations. This is particularly relevant in $(d \gg N)$ regime, where optimization may otherwise commit early to suboptimal sparse configurations with poor generalization.

We formulate a differentiable, entropy-regularized, $L_0$-constrained objective using Hard Concrete gates, and show how to optimize it in a federated setting using reparameterized gradients at clients and aggregation of sparse updates. The resulting method helps achieve a user-defined target sparsity while improving exploration of the parameter space during learning.

Experiments on synthetic linear regression, CNN for image classification and softmax classifier for leukemia classification show that \texttt{E-FLoPS} consistently improves sparsity recovery and statistical performance at the target sparsity compared with iterative hard-thresholding-based sparse federated training, \texttt{Fed-IHT}, and post-training pruning of dense \texttt{FedAvg}. These gains are achieved with communication efficiency comparable to \texttt{Fed-IHT}.

We show that \texttt{E-FLoPS} is effective for both linear and non-linear models at high sparsity levels. Understanding how entropy regularization affects sparse recovery differently in data-scarce versus overparameterized regimes is an interesting direction for future work, along with extending the method to structured sparsity.
\begin{acknowledgments}
This work has been supported by FAST, the Finnish Software Engineering Doctoral Research Network, funded by the Ministry of Education and Culture in Finland.
\end{acknowledgments}

\section*{Declaration on Generative AI}
During the preparation of this work, the author(s) used Grammarly in order to: Grammar and
spelling check in Overleaf.
After using these tool(s)/service(s), the author(s) reviewed and edited the content as needed and take(s)
full responsibility for the publication’s content.

\bibliographystyle{plainnat}
\bibliography{bibliography}

@inproceedings{1,
  title={Communication-efficient learning of deep networks from decentralized data},
  author={McMahan, Brendan and Moore, Eider and Ramage, Daniel and Hampson, Seth and y Arcas, Blaise Aguera},
  booktitle={Artificial intelligence and statistics},
  pages={1273--1282},
  year={2017},
  organization={PMLR}
}

@article{2,
  title={Advances and open problems in federated learning},
  author={Kairouz, Peter and McMahan, H Brendan and Avent, Brendan and Bellet, Aur{\'e}lien and Bennis, Mehdi and Bhagoji, Arjun Nitin and Bonawitz, Kallista and Charles, Zachary and Cormode, Graham and Cummings, Rachel and others},
  journal={Foundations and trends{\textregistered} in machine learning},
  volume={14},
  number={1--2},
  pages={1--210},
  year={2021},
  publisher={Now Publishers, Inc.}
}

@article{tibshirani1996regression,
  title={Regression shrinkage and selection via the lasso},
  author={Tibshirani, Robert},
  journal={Journal of the Royal Statistical Society Series B: Statistical Methodology},
  volume={58},
  number={1},
  pages={267--288},
  year={1996},
  publisher={Oxford University Press}
}

@article{Louizos2017LearningSN,
  title={Learning sparse neural networks through $ L\_0 $ regularization},
  author={Louizos, Christos and Welling, Max and Kingma, Diederik P},
  journal={arXiv preprint arXiv:1712.01312},
  year={2017}
}

@article{GallegoPosada2022ControlledSV,
  title={Controlled sparsity via constrained optimization or: How i learned to stop tuning penalties and love constraints},
  author={Gallego-Posada, Jose and Ramirez, Juan and Erraqabi, Akram and Bengio, Yoshua and Lacoste-Julien, Simon},
  journal={Advances in Neural Information Processing Systems},
  volume={35},
  pages={1253--1266},
  year={2022}
}

@article{Maddison2016TheCD,
  title={The concrete distribution: A continuous relaxation of discrete random variables},
  author={Maddison, Chris J and Mnih, Andriy and Teh, Yee Whye},
  journal={arXiv preprint arXiv:1611.00712},
  year={2016}
}

@article{bertsimas_sparse_2020,
 ISSN = {08834237, 21688745},
 URL = {https://www.jstor.org/stable/26997931},
 author = {Dimitris Bertsimas and Jean Pauphilet and Bart Van Parys},
 journal = {Statistical Science},
 number = {4},
 pages = {pp. 555--578},
 publisher = {Institute of Mathematical Statistics},
 title = {Sparse Regression: Scalable Algorithms and Empirical Performance},
 urldate = {2025-09-05},
 volume = {35},
 year = {2020}
}

@article{tong_federated_2022,
  title={Federated Optimization of L0-norm Regularized Sparse Learning},
  author={Tong, Qianqian and Liang, Guannan and Ding, Jiahao and Zhu, Tan and Pan, Miao and Bi, Jinbo},
  journal={Algorithms},
  volume={15},
  number={9},
  pages={319},
  year={2022},
  publisher={MDPI}
}

@article{solans2024non,
  title={Non-IID data in federated learning: A survey with taxonomy, metrics, methods, frameworks and future directions},
  author={Solans, David and Heikkila, Mikko and Vitaletti, Andrea and Kourtellis, Nicolas and Anagnostopoulos, Aris and Chatzigiannakis, Ioannis and others},
  journal={arXiv preprint arXiv:2411.12377},
  year={2024}
}

@article{reisizadeh2020robust,
  title={Robust federated learning: The case of affine distribution shifts},
  author={Reisizadeh, Amirhossein and Farnia, Farzan and Pedarsani, Ramtin and Jadbabaie, Ali},
  journal={Advances in neural information processing systems},
  volume={33},
  pages={21554--21565},
  year={2020}
}

@article{lecun1998mnist,
  title={The MNIST database of handwritten digits},
  author={LeCun, Yann},
  journal={http://yann. lecun. com/exdb/mnist/},
  year={1998}
}

@phdthesis{ranganath2017black,
  title={Black Box variational inference: Scalable, generic Bayesian computation and its applications},
  author={Ranganath, Rajesh},
  year={2017},
  school={Princeton University}
}

@inproceedings{horn1990hadamard,
  title={The hadamard product},
  author={Horn, Roger A},
  booktitle={Proc. symp. appl. math},
  volume={40},
  pages={87--169},
  year={1990}
}

@misc{wang2021fieldguidefederatedoptimization,
      title={A Field Guide to Federated Optimization}, 
      author={Jianyu Wang and Zachary Charles and Zheng Xu and Gauri Joshi and H. Brendan McMahan and Blaise Aguera y Arcas and Maruan Al-Shedivat and Galen Andrew and Salman Avestimehr and Katharine Daly and Deepesh Data and Suhas Diggavi and Hubert Eichner and Advait Gadhikar and Zachary Garrett and Antonious M. Girgis and Filip Hanzely and Andrew Hard and Chaoyang He and Samuel Horvath and Zhouyuan Huo and Alex Ingerman and Martin Jaggi and Tara Javidi and Peter Kairouz and Satyen Kale and Sai Praneeth Karimireddy and Jakub Konecny and Sanmi Koyejo and Tian Li and Luyang Liu and Mehryar Mohri and Hang Qi and Sashank J. Reddi and Peter Richtarik and Karan Singhal and Virginia Smith and Mahdi Soltanolkotabi and Weikang Song and Ananda Theertha Suresh and Sebastian U. Stich and Ameet Talwalkar and Hongyi Wang and Blake Woodworth and Shanshan Wu and Felix X. Yu and Honglin Yuan and Manzil Zaheer and Mi Zhang and Tong Zhang and Chunxiang Zheng and Chen Zhu and Wennan Zhu},
      year={2021},
      eprint={2107.06917},
      archivePrefix={arXiv},
      primaryClass={cs.LG},
      url={https://arxiv.org/abs/2107.06917}, 
}

@article{ent1,
  title={Approximate sparsity pattern recovery: Information-theoretic lower bounds},
  author={Reeves, Galen and Gastpar, Michael C},
  journal={IEEE Transactions on Information Theory},
  volume={59},
  number={6},
  pages={3451--3465},
  year={2013},
  publisher={IEEE}
}

@article{ent2,
  title={Necessary and sufficient conditions for sparsity pattern recovery},
  author={Fletcher, Alyson K and Rangan, Sundeep and Goyal, Vivek K},
  journal={IEEE Transactions on Information Theory},
  volume={55},
  number={12},
  pages={5758--5772},
  year={2009},
  publisher={IEEE}
}

@article{ent3,
  title={Recovery from compressed measurements using sparsity independent regularized pursuit},
  author={Thomas, Thomas James and Rani, J Sheeba},
  journal={Signal Processing},
  volume={172},
  pages={107508},
  year={2020},
  publisher={Elsevier}
}

@article{ent4,
  title={The sampling rate-distortion tradeoff for sparsity pattern recovery in compressed sensing},
  author={Reeves, Galen and Gastpar, Michael},
  journal={IEEE Transactions on Information Theory},
  volume={58},
  number={5},
  pages={3065--3092},
  year={2012},
  publisher={IEEE}
}

@inproceedings{ent5,
  title={Maximum entropy information bottleneck for uncertainty-aware stochastic embedding},
  author={An, Sungtae and Jammalamadaka, Nataraj and Chong, Eunji},
  booktitle={Proceedings of the IEEE/CVF Conference on Computer Vision and Pattern Recognition},
  pages={3809--3818},
  year={2023}
}

@article{ent6,
  title={Understanding disentangling in $\beta$-VAE},
  author={Burgess, Christopher P and Higgins, Irina and Pal, Arka and Matthey, Loic and Watters, Nick and Desjardins, Guillaume and Lerchner, Alexander},
  journal={arXiv preprint arXiv:1804.03599},
  year={2018}
}

@inproceedings{ent7,
  title={Sparse Bayesian networks: efficient uncertainty quantification in medical image analysis},
  author={Abboud, Zeinab and Lombaert, Herve and Kadoury, Samuel},
  booktitle={International Conference on Medical Image Computing and Computer-Assisted Intervention},
  pages={675--684},
  year={2024},
  organization={Springer}
}

@article{ent8,
  title={Federated learning via variational bayesian inference: Personalization, sparsity and clustering},
  author={Zhang, Xu and Li, Wenpeng and Shao, Yunfeng and Li, Yinchuan},
  journal={arXiv preprint arXiv:2303.04345},
  year={2023}
}

@misc{huthasana2025federatedlearningl0constraint,
      title={Federated Learning With L0 Constraint Via Probabilistic Gates For Sparsity}, 
      author={Krishna Harsha Kovelakuntla Huthasana and Alireza Olama and Andreas Lundell},
      year={2025},
      eprint={2512.23071},
      archivePrefix={arXiv},
      primaryClass={stat.ML},
      url={https://arxiv.org/abs/2512.23071}, 
}

@article{golub1999molecular,
  title={Molecular classification of cancer: class discovery and class prediction by gene expression monitoring},
  author={Golub, Todd R and Slonim, Donna K and Tamayo, Pablo and Huard, Christine and Gaasenbeek, Michelle and Mesirov, Jill P and Coller, Hilary and Loh, Mignon L and Downing, James R and Caligiuri, Mark A and others},
  journal={science},
  volume={286},
  number={5439},
  pages={531--537},
  year={1999},
  publisher={American Association for the Advancement of Science}
}

@article{liang2013sparse,
  title={Sparse logistic regression with a L1/2 penalty for gene selection in cancer classification},
  author={Liang, Yong and Liu, Cheng and Luan, Xin-Ze and Leung, Kwong-Sak and Chan, Tak-Ming and Xu, Zong-Ben and Zhang, Hai},
  journal={BMC bioinformatics},
  volume={14},
  number={1},
  pages={198},
  year={2013},
  publisher={Springer}
}

@inproceedings{bao2022fast,
  title={Fast composite optimization and statistical recovery in federated learning},
  author={Bao, Yajie and Crawshaw, Michael and Luo, Shan and Liu, Mingrui},
  booktitle={international conference on machine learning},
  pages={1508--1536},
  year={2022},
  organization={PMLR}
}

\appendix
\section{KL-Divergence for Hard Concrete Gates}\label{C}

The stochastic gates follow the Hard Concrete distribution \citep{Louizos2017LearningSN}, obtained by stretching a Binary Concrete random variable $s \in (0,1)$:
\begin{equation}
\bar{s} = s(\zeta - \gamma) + \gamma, \quad
z = \min(1, \max(0,\bar{s})),
\end{equation}
with $\gamma < 0$ and $\zeta > 1$. This induces a mixed distribution over $z$ with point masses at $0$ and $1$, and a continuous component on $(0,1)$:
\begin{align}
q(z \mid \phi)
= Q_{\bar{s}}(0)\,\delta(z)
+ \left(1 - Q_{\bar{s}}(1)\right)\,\delta(z-1)
+ \left(Q_{\bar{s}}(1) - Q_{\bar{s}}(0)\right)
q_{\bar{s}}(z \mid \bar{s}\in(0,1)),
\end{align}
\begin{equation}
\text{where}, \quad Q_{\bar{s}}(\bar{s} \mid \phi) =
Q_s\left(
\frac{\bar{s} - \gamma}{\zeta - \gamma}
\,\middle|\, \phi
\right) \quad \text{and,}\quad Q_s(s \mid \phi) =
\sigma\left(
\beta' (\log s - \log(1-s)) - \phi
\right). 
\end{equation}
we define a hard concrete prior $p(z)$ with the same support and then the KL-divergence decomposes as:
\begin{align}
\mathrm{KL}(q(z)\|p(z))
&= Q_{\bar{s}}(0)\log\frac{Q_{\bar{s}}(0)}{P_{\bar{s}}(0)}
+ \left(1-Q_{\bar{s}}(1)\right)
\log\frac{1-Q_{\bar{s}}(1)}{1-P_{\bar{s}}(1)} \\
&\quad + \left(Q_{\bar{s}}(1)-Q_{\bar{s}}(0)\right)
\mathbb{E}_{q_{\bar{s}}(z|\bar{s}\in(0,1))}\big[\log q_{\bar{s}}(z) - \log p_{\bar{s}}(z)\big].
\end{align}

The first two terms correspond to masses at $z=0$ and $z=1$, and the last term accounts for the continuous component. In practice, this divergence is evaluated either using the closed-form expressions if available or Monte Carlo estimation. We used the closed form expressions provided at \citet{Louizos2017LearningSN}.
\end{document}